\documentclass[lettersize,journal]{IEEEtran}
\usepackage{amsmath,amsfonts}
\usepackage{algpseudocode}
\usepackage{algorithm}
\usepackage{array}
\usepackage[caption=false,font=normalsize,labelfont=sf,textfont=sf]{subfig}
\usepackage{textcomp}
\usepackage{stfloats}
\usepackage{url}
\usepackage{verbatim}
\usepackage{graphicx}
\usepackage{orcidlink}
\usepackage{cite}
\begin{document}

\title{Improve Value Estimation of Q Function and Reshape Reward with Monte Carlo Tree Search}

\author{Jiamian Li~\orcidlink{0009-0005-6713-2707}}



\maketitle

\begin{abstract}
Reinforcement learning has achieved remarkable success in perfect information games such as Go and Atari, enabling agents to compete at the highest levels against human players. However, research in reinforcement learning for imperfect information games has been relatively limited due to the more complex game structures and randomness. Traditional methods face challenges in training and improving performance in imperfect information games due to issues like inaccurate Q value estimation and reward sparsity. In this paper, we focus on Uno, an imperfect information game, and aim to address these problems by reducing Q value overestimation and reshaping reward function. We propose a novel algorithm that utilizes Monte Carlo Tree Search to average the value estimations in Q function. Even though we choose Double Deep Q Learning as the foundational framework in this paper, our method can be generalized and used in any algorithm which needs Q value estimation, such as the Actor-Critic. Additionally, we employ Monte Carlo Tree Search to reshape the reward structure in the game environment. We compare our algorithm with several traditional methods applied to games such as Double Deep Q Learning, Deep Monte Carlo and Neural Fictitious Self Play, and the experiments demonstrate that our algorithm consistently outperforms these approaches, especially as the number of players in Uno increases, indicating a higher level of difficulty.

\end{abstract}


\section{Introduction}
\IEEEPARstart{I}{n} previous research, there has been considerable exploration into the application of reinforcement learning in perfect information games, such as Go \cite{silver2017mastering} \cite{silver2016mastering}, Chess\cite{silver2017masteringchess} \cite{silver2018general} and Atari\cite{mnih2013playing} \cite{van2016deep}. These studies have demonstrated significant success, with reinforcement learning agents consistently outperforming top human players through the deep neural networks to derive optimal policies. However, research in the domain of imperfect information games has been relatively scarce, and significant breakthroughs in this area remain limited. Several factors contribute to this difference. 
Imperfect information games are defined by the fact that players cannot access all the information within a given state, leading to varying levels of knowledge among participants, unlike in perfect information games. For instance, in most card games, players can only view their own hands, while crucial details, such as the cards held by opponents and those remaining in the deck, remain unknown.
Additionally, these games are inherently non-deterministic and often include a high degree of randomness and luck, which presents significant challenges for reinforcement learning algorithms during training, potentially leading to convergence issues or high variance in results. 
Reinforcement learning agents in imperfect information games also encounter the challenge of sparse rewards, where obtaining rewards can be difficult, and positive feedback may be delayed for extended periods. This increases the computational demand for sampling and renders much of the sampled data ineffective.
In previous research on imperfect information games like StarCraft\cite{vinyals2017starcraft}\cite{vinyals2019grandmaster}\cite{liu2021efficient}, Doudizhu \cite{zha2021douzero} \cite{zhao2022douzero+} \cite{zhao2023full} and  Texas Holdem \cite{heinrich2016deep} \cite{brown2019deep}, Deep Monte Carlo(DMC)\cite{sutton2018reinforcement}, Q-learning\cite{watkins1992q}\cite{van2016deep} and Neural Fictitious Self Play(NFSP)\cite{heinrich2016deep}\cite{zhang2021monte} have been the dominant and commonly used methods, yielding some results; however, they suffer from the issues such as overestimation of Q values\cite{van2016deep}, inefficient for incomplete episodes and large variance of expected returns (rewards)\cite{greensmith2004variance}, and often require an extensive number of training episodes to achieve satisfactory performance. These challenges highlight the necessity for new methods to improve estimation of values of a state, enhance convergence speed and increase sample efficiency.

We choose an imperfect information game, Uno, as the training environment for our reinforcement learning agent in this paper. Uno\cite{Unorules} is a popular game played worldwide, suitable for 2 to 10 players. The game consists of multiple card types in four colors: red, yellow, blue, and green. The numbered cards range from 0 to 9, and there are special cards such as Skip, Reverse, Draw 2, Wild, and Wild Draw 4. Each color contains two of each number card from 1 to 9, and one card with the number 0. Additionally, each color has two Skip, Reverse, and Draw 2 cards. There are four Wild cards and four Wild Draw 4 cards, making a total of 108 cards. Table \ref{uno_rule} provides details about the function cards, while example cards are shown in Figure \ref{tab:uno}. The rules of Uno are as follows:
\begin{enumerate}
    \item \textbf{Initialize:} Shuffle the Uno deck, deal 7 cards to each player, and place the remaining cards in the center as the draw pile.
    
    \item \textbf{Start:} Flip over the top card of the draw pile to form the target card. Players must match either the color or number of this target card, or play a special card that fits these rules, such as Wild or Wild Draw 4.
    
    \item \textbf{Play:} If a player has a valid card, they must play it, setting it as the new target card, and the old target card is discarded. If they cannot play a valid card, they must draw a card from the draw pile. If the draw pile runs out, shuffle the discarded cards to create a new draw pile.
    
    \item \textbf{Win:} The first player to get rid of all their cards wins the game. If a player is left with only one card in hand, they must call out "Uno." Failure to do so results in a penalty, and the player must draw 2 additional cards.
\end{enumerate}

\begin{table*}[h!]
    \caption{Functional Description of Special Cards\label{tab:uno}}
    \label{uno_rule}
    \centering
    \begin{tabular}{c c} 
        \hline
        \hline
        \textbf{Type of Card} & \textbf{Card Function} \\ 
        \hline
        Skip & Skip the next player's turn, allowing the player after them to take their turn instead. \\ 
        \hline
        Reverse & Reverse the direction of play: switch between clockwise and counterclockwise. \\ 
        \hline
        Draw 2 & Make the next player draw two cards. \\ 
        \hline
        Wild & Can be played as any color. \\ 
        \hline
        Wild Draw 4 & Can be played as any color and make the next player draw 4 cards. \\ 
        \hline
    \end{tabular}
\end{table*}

\begin{figure}[!htbp]
\centering
\includegraphics[width=2 in]{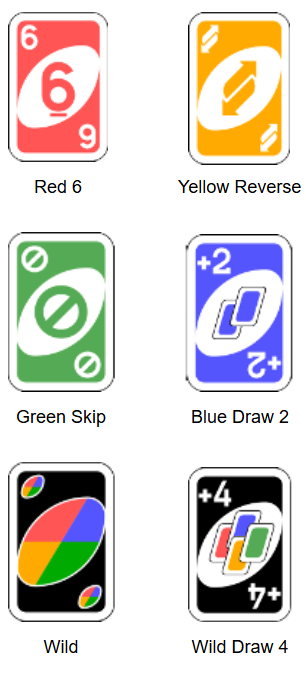}
\caption{The different kind of Uno cards including the
Number card, Skip, Reverse, Draw 2, Wild and Wild Draw 4.}
\label{uno}
\end{figure}
\begin{figure}[!htbp]
\centering
\includegraphics[width=2.5 in]{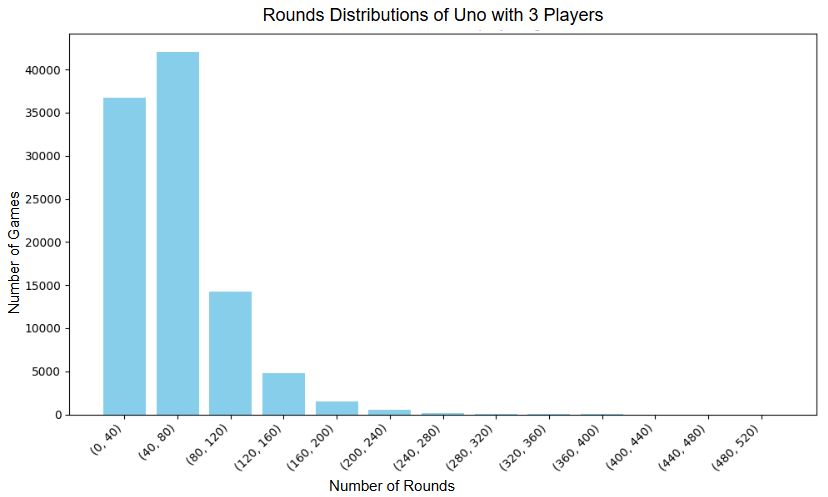}
\caption{The round distribution in the Uno game with 3
players. We record rounds of one hundred thousand games in total. Most games typically end within around 40 - 80 rounds, but there
are some games that can extend beyond 200 rounds. }
\label{round}
\end{figure}

Uno, like other imperfect information games, suffers from the issue of reward sparsity, but to a more severe degree. The Uno rules lead to situations where the deck is exhausted, requiring a reshuffle, until one player plays all their cards and win. This results in more rounds in Uno compared to other games (round distribution shown in Figure \ref{round}).

In this paper, we aim to address the issues of reward sparsity and convergence challenges associated with problems of inaccurate estimation of Q values. To achieve this, we propose a new algorithm that improves value estimation of Q functions using Monte Carlo Tree Search (MCTS)\cite{browne2012survey} to solve Uno. Additionally, we utilize MCTS to reshape the reward structure within the Uno environment. We choose Double Deep Q-Learning (DDQN) as the base algorithm for training the agent and extend it to DDQN with MCTS; however, our method can be applied to any algorithm that requires Q value estimation. Our algorithm is trained for the same number of steps as DMC, NFSP, and DDQN in environments with three and four players. The evaluation results indicate that DDQN with MCTS achieves a higher win rate than the other algorithms by 4\% - 16\%. More importantly, the integration of MCTS into DDQN exhibits a marked acceleration in performance improvement during the early and mid-training phases, whereas other algorithms may either show no significant progress or improve at a much slower rate.

\section{Related Work}

\subsection{Partially Observable Markov Decision Process}
\begin{figure}[!htbp]
\centering
\includegraphics[width=2.5 in]{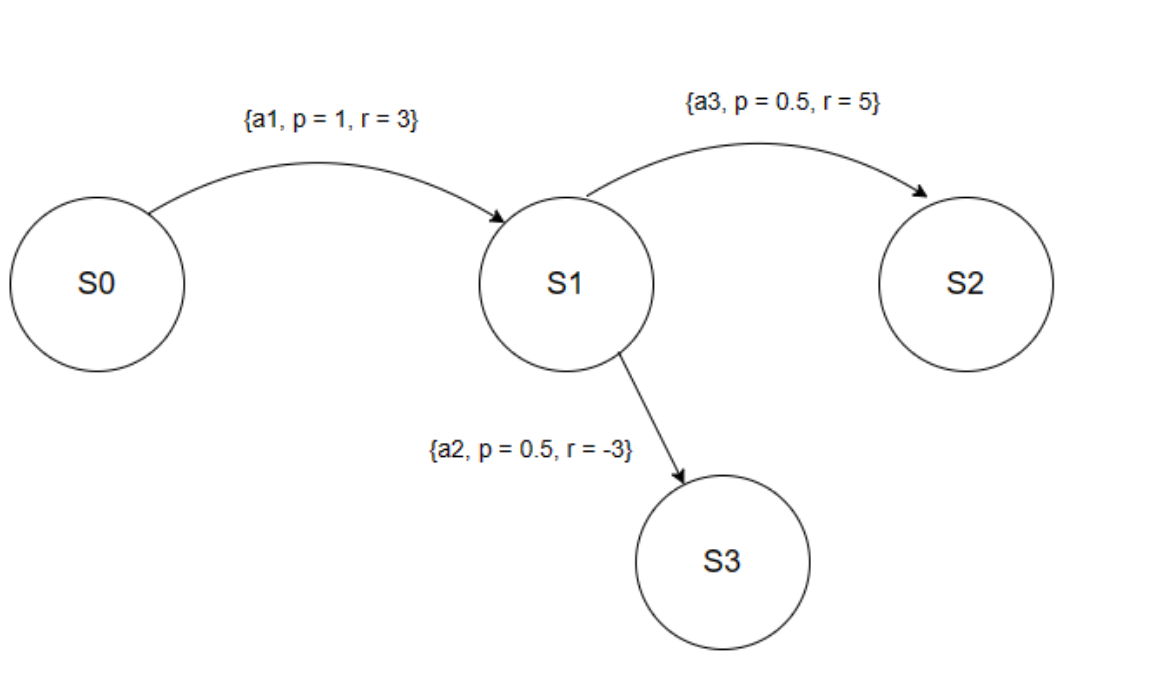}
\caption{The simplified Markov Decision Process (MDP) consists of four states, $S_0 - S_3$, which represent the possible states. The actions, $a_1 - a_3$, denote the decisions the agent makes to transition from one state to another. The transition function, $p$, represents the probability (greater than or equal to 0) of the agent taking a particular action at a state. The reward function, $r$, which can be either positive or negative, defines the immediate reward the agent receives upon transitioning to a new state.}
\label{MDP}
\end{figure}

Any classic reinforcement learning problems can be modelled as a mathematical framework called Markov Decision Process (MDP)\cite{sutton2018reinforcement}, shown in Figure \ref{MDP}. In academic formalism, the MDP can be described as a five-tuple $(S, A, T, R, \lambda)$, where:

\begin{itemize}
    \item $S$ is the set of all possible states in the environment.
    \item $A$ is the set of all possible actions the agent can take in the environment.
    \item $T(s'|s,a)$ is the state transition probability function, which is the probability of transitioning to the next state $s'$ when agents take action $a$ in state $s$.
    \item $R(s,a,s')$ is the reward function, which is the immediate reward received when transitioning from state $s$ to next state $s'$ by taking action $a$.
    \item $\lambda$ is the discount factor, which determines how future rewards are valued compared to immediate rewards, typically ranging from $0$ to $1$.
\end{itemize}

MDP assumes that all information about a state in the environment is accessible to agents. However, imperfect information games do not meet this condition, so they are modeled as a broader framework: Partially Observable Markov Decision Process (POMDP) \cite{spaan2012partially}. POMDP can be described as a seven-tuple $(S, A, T, R, \Omega, O, \lambda)$, where $(S, A, T, R, \lambda)$ are the same as in MDP, while the additional two elements can be defined as:

\begin{itemize}
    \item $\Omega$ is the set of all possible observations that the agent can perceive.
    \item $O(\omega'|s',a)$ is the observation probability function, which specifies the probability of observing $\omega$ when transitioning to state $s'$ after taking action $a$.
\end{itemize}

In POMDP, the agent, based on the current state $s$ and observations $\omega$, selects an action $a$, transitions to a new state $s'$ according to the state transition probability $P$, receives a new observation based on $O(\omega'|s',a)$, and obtains a reward $R(s, a, s')$. To solve POMDP, agents need to find an optimal policy $\pi^*$ that maximizes the expected cumulative reward over time. The optimal policy $\pi^*$ can be either deterministic or stochastic, selecting the best action given the current state and observations.

\subsection{Q-learning and Deep Q-network}
Q-learning\cite{sutton2018reinforcement} is one of the foundational and most widely-used algorithms in reinforcement learning, playing a critical role as both a core component and an essential building block for more advanced algorithms across various domains. One of the key characteristics of Q-learning is its reliance on the Bellman optimal equation for updates. With sufficient training data, Q-learning can guarantee that the agent converges to the optimal policy. As a model-free, bootstrapping algorithm, Q-learning learns to estimate the optimal action to take in any given state by calculating and updating Q values—representing the expected utility of taking a particular action from a specific state. Q-learning operates on a Q-table where each state-action pair is associated with a Q value. The algorithm updates its Q values through the following steps in each iteration:

\begin{enumerate}
    \item \textbf{Select and execute an action}: Select an action according to the current policy, such as the $\epsilon$-greedy strategy, which typically selects the best current action (in most cases) but with a very small probability randomly selects another action to explore unknown state-actions.
    \item \textbf{Observe the outcome and reward}: After executing the action, agents observe the new state and the reward received.
    \item \textbf{Update the Q value}: Update the Q-table using the following update rule:
    \[
    Q(s, a) \leftarrow Q(s, a) + \alpha \left( r + \lambda \max_{a'} Q(s', a') - Q(s, a) \right)
    \]
    where $s$ and $a$ are the original state and action taken, $s'$ is the next state after taking the action, $r$ is the immediate reward received, $\alpha$ is the learning rate, and $\lambda$ is the discount factor, which determines the decay rate of the future reward.
\end{enumerate}

This iterative process allows the agent to learn an optimal strategy over time by refining its estimate of the Q values for each state-action pair, ultimately converging on the policy that maximizes cumulative rewards.

Q-learning is well-suited for problems with small size of state and action spaces. However, in large-scale spaces, the traditional Q-learning method becomes impractical, as the Q-table can grow excessively large, making it inefficient to learn and store. To address this, Q-learning was replaced by Deep Q-Network(DQN) \cite{mnih2015human}, which employs neural networks as Q value approximators to overcome the challenges of large state and action spaces. In addition, DQN introduced the technique of experience replay, where the agent's experiences are stored in memory and used for batch updates rather than updating immediately after each transition. This approach increases sample efficiency by allowing the agent to learn from past experiences multiple times, improving the use of available data. Although DQN has become a benchmark algorithm in reinforcement learning, it has some drawbacks, one of the main issues being its tendency to overestimate Q values. To address this, Double Deep Q-Network (DDQN) \cite{van2016deep} was introduced. DDQN tackles the overestimation by using two separate neural networks: one network (estimator network) is used to predict the current Q values for action selection, while the other network (target network), which is a past version of the estimator network, is used to calculate the target Q values for updating. By separating these two processes and using a target network that lags behind the estimator, DDQN reduces the bias in estimating the target values, leading to more accurate and stable learning. It tries to minimize the loss between target value and estimated Q value to train the agent:

\begin{align}
L_{\text{ddqn}} =  (Q_{\text{target}}(s, a; \theta') - Q_{\text{estimator}}(s, a; \theta))^2,
\label{eq_ddqn1}
\end{align}
where 
\begin{align}
Q_{\text{target}}(s, a; \theta') &= r + \lambda Q_{\text{target}} \left(s', \arg\max_{a'} Q_{\text{estimator}}(s', a'; \theta); \theta'\right)
\label{eq_ddqn2}
\end{align}
$Q_{\text{target}}$ and $Q_{\text{estimator}}$ are the target network and the estimator network, respectively.

\subsection{Deep Monte Carlo and Monte Carlo Tree Search}
Deep Monte Carlo (DMC) is similar to DQN; however, while DQN is based on bootstrapping, DMC relies on complete trajectories. In DQN, updates are made based on estimated future returns, whereas DMC calculates the true return from the entire episode, based on Monte Carlo sampling. Although the DMC method is known for its high variance\cite{sutton2018reinforcement}, it can be effective in certain episodic games, such as the card game DouDiZhu\cite{zha2021douzero}. DMC agents usually selects a random policy $\pi$ at the start and  optimizes $\pi$ through the following steps:
\begin{enumerate}
    \item Generate an episode using $\pi$ and store the tuple $(s_t, a_t, R(s_t))$ for each step.
    \item Initialize the return $G(s)$ of each state $s$ at time $t$ to 0. The average return for each $s$ is calculated using the formula $G(s_t) = R(s_t) + \lambda G(s_{t+1})$, where $\lambda$ is the reward discount factor.
    \item Minimize the mean squared error between $G(s_t)$ and $Q(s_t, a_t)$, where $Q(s_t, a_t)$ is predicted by the deep neural network. Repeat steps from 1 - 3 and finally, the optimal policy is generated by updating: For each $s$ in the episode, $\pi(s) \leftarrow \arg\max_a Q(s, a)$.
\end{enumerate}
Another method based on Monte Carlo sampling is Monte Carlo Tree Search (MCTS), but unlike DMC, it builds a search tree by iterating over simulations of possible future moves and backpropagating the results, which makes it a key component of the renowned Go agent AlphaGo's\cite{silver2016mastering} algorithm. The MCTS process in reinforcement learning includes four principal stages:
\begin{enumerate}
    \item \textbf{Selection}: Starting from the root node, repeatedly select child nodes based on a combination of predicted state values from the neural network and the Upper Confidence Bound for Trees (UCT) formula.
    \item \textbf{Expansion}: If the selected node represents a non-terminal state and has not been fully expanded, add one of its unvisited child nodes to the tree and use the neural network to predict the value of this new state.
    \item \textbf{Simulation}: From the newly expanded node, simulate the game by following a policy until a terminal state is reached, where the outcome of the game is determined.
    \item \textbf{Backpropagation}: Once the simulation is complete, propagate the result back through the nodes along the path to the root. Update the visit counts at each node based on the outcome of the simulation.
\end{enumerate}

\subsection{Neural Fictitious Self Play}
Many reinforcement learning algorithms on multi-players' games are based on Nash equilibrium\cite{kreps1989nash}, with Neural Fictitious Self-Play (NFSP)\cite{heinrich2016deep} being a notable example. Nash equilibrium is a crucial concept in game theory, especially in non-cooperative games. It describes a situation where multiple participants select their optimal strategies, assuming that all other players will keep their strategies unchanged. In this equilibrium, no player can improve their payoff by unilaterally altering their strategy. 
In the mathematical form, a Nash equilibrium of a game with two players can be expressed as:
\[
u_1(s) \geq \max_{s_1 \in S_1} u_1(s_1, s_2)
\]

\[
u_2(s) \geq \max_{s_2 \in S_2} u_2(s_1, s_2),
\]
where $u_1$, $u_2$ are the payoffs (reward) of player 1 and player 2, and $s$ is the strategy. NFSP employs neural networks to approximate Nash equilibrium by responding to opponents' average strategies. It builds on game-theoretical principles by utilizing two parallel learning processes: one focuses on learning the average policy across all agents, while the other estimates the value of the policy, typically implemented through a Q-learning-based approach. NFSP is composed of two core networks: a value network, the Deep Q-Network $(Q(s, a|\theta_Q))$, which estimates action values, and a policy network $\Pi(s, a|\theta_\Pi)$, responsible for mimicking the agent's past best responses. Both networks are supported by corresponding memory buffers: Memory for Reinforcement Learning(MRL) stores experiences for Q-learning, while Memory for Supervised Learning(MSL) stores the best-response behaviors. The agent's learning process is a combination of data drawn from both networks. Experiences in MRL are used to train the value network by minimizing the mean squared error between predicted values and the stored experiences in the replay buffer. At the same time, data from MSL is used to train the policy network by minimizing the negative log-likelihood between the stored actions and those predicted by the policy network. NFSP has proven to be highly effective, particularly in complex imperfect-information games like Leduc Hold'em and Limit Hold'em. At the time of its development, it demonstrated near-superhuman performance in these environments, outperforming state-of-the-art algorithms and setting a new benchmark in imperfect information game strategies.

\section{METHODOLOGY}
\subsection{Uno Environment and Representations in Reinforcement Learning}
We use RLCard \cite{zha2019rlcard}, which is the Python game toolkit, as our Uno framework and reinforcement learning environment. One problem for our method is how to represent the game into a reinforcement learning environment. We have to define the elements of the MDP in the context of UNO:

\begin{itemize}
    \item \textbf{State (S)}: The state represents the current situation of a single player, which only includes information available for that player. In our state encoding of UNO, this is characterized by the player’s hand and the target card. Different players would have different states as they have different hands.
    \item \textbf{Actions (A)}: Actions are the legal moves that a player can make during their turn in a round of UNO.
    \item \textbf{Transition Function (T)}: The transition function defines the probability distribution of legal moves given a player's current state, which can be predicted by neural networks, which is $\pi(a|s, \theta)$.
    \item \textbf{Reward (R)}: The reward is what a player gains after completing a round. In the basic rule of UNO, a reward of +1 is granted exclusively to the winner who has no cards left, while all others receive -1 at the end. During the ongoing game, the reward following all actions is 0. But our algorithm will reshape the reward structure based on MCTS, allowing the agent to receive rewards for certain actions taken.
    \item \textbf{Discount Factor ($\lambda$)}: This is a hyperparameter signifying the degree to which future rewards are considered relative to immediate ones. For the purpose of our analysis, we have set this value at 0.99.
\end{itemize}

It is essential to abstract states and actions into a format suitable for neural networks based on our definitions. In Uno, there are 61 distinct types of actions, which defines the action space size as 61. Each action can be represented by a unique integer ranging from 0 to 60, the same encoding approach used in RLCard. Our neural networks will produce an output vector of size 61, with each element representing the Q-value associated with its corresponding action. Detailed action encodings are shown in Table~\ref{tab:card_encoding}.

\begin{table}[!h]
    \centering
    \caption{Action Encoding of Uno}
    \label{tab:card_encoding}
    \begin{tabular}{|c|c|}
        \hline
        \textbf{Action ID} & \textbf{Card Info} \\
        \hline
        0-9 & Red Cards with Numbers \\
        10 & Red Skip \\
        11 & Red Reverse \\
        12 & Red Draw 2 \\
        13 & Red Wild \\
        14 & Red Wild 4 \\
        15-24 & Green Cards with Numbers \\
        25 & Green Skip \\
        26 & Green Reverse \\
        27 & Green Draw 2 \\
        28 & Green Wild \\
        29 & Green Wild 4 \\
        30-39 & Blue Cards with Numbers \\
        40 & Blue Skip \\
        41 & Blue Reverse \\
        42 & Blue Draw 2 \\
        43 & Blue Wild \\
        44 & Blue Wild 4 \\
        45-54 & Yellow Cards with Numbers \\
        55 & Yellow Skip \\
        56 & Yellow Reverse \\
        57 & Yellow Draw 2 \\
        58 & Yellow Wild \\
        59 & Yellow Wild 4 \\
        60 & Draw \\
        \hline
    \end{tabular}
\end{table}
We have adopted the state encoding approach described in RLCard, but our method differs by reducing the state size and omitting unnecessary information. While RLCard includes the player’s hand, the target card, and additional cards as part of the state, we argue that incorporating additional cards is redundant. Cards outside a player’s hand and target card include opponents' cards and the deck, which is reshuffled in UNO. Since the agent can't distinguish them in neural networks, this adds unnecessary complexity and may hinder training. Including additional cards would also vastly expand the state space. To simplify and speed up training, we use only the player's hand and target card. The state size of RLCard's state encoding is $10^{126}$, while ours is $10^{72}$.

We encode hands and the target card into 4 planes. Every plane is a matrix of size 4x15, with each entry being 0 or 1 for one-hot encodings. The number 15 represents the different types of cards, disregarding color, which are the number cards from 0-9 and five kinds of special cards: skip, reverse, draw 2, wild, and wild 4. The number 4 represents the four colors: yellow, green, blue and red. 

\begin{figure}[!htbp]
\centering
\includegraphics[width=3 in]{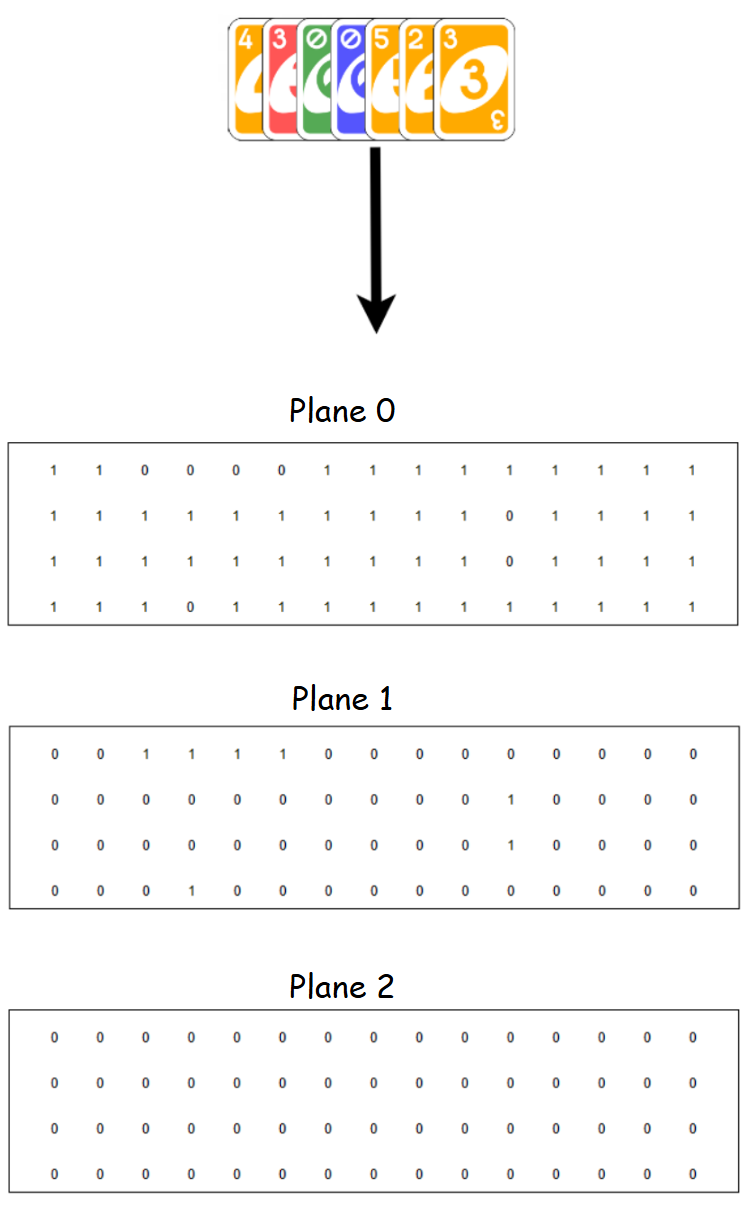}
\caption{The exampled hand encodings. In each plane, the first four rows represent the colors yellow, green, blue, and red, while the columns correspond to the number cards from 0 to 9, as well as the action cards: skip, reverse, draw 2, wild, and wild 4.}
\label{state}
\end{figure}

The agent's hand information is encoded into three planes(shown in Figure~\ref{state}), as in Uno, a player can have either 0, 1, or at most 2 cards of any given type. The plane 0 indicates that the player has zero of such cards in their hand. The plane 1 indicates that the player has exactly one of such cards. The plane 2 indicates that the player has two of such cards. For example, if the player has no Red 8 cards, the entry for Red 8, located in the fourth row and the ninth column, in the plane 0 is 1, while the entries for Red 8 in all other planes are 0. If the player has two Red 8 cards, the entry for Red 8 in the plane 2 is 1, and the entries for Red 8 in all other planes are 0. We encode the target card into a single plane where the matrix entry corresponding to the target card is 1, and the values of all other entries in the matrix are 0. 

\subsection{Improve Q value estimation in DDQN and reshape rewards with Monte Carlo Tree Search}
Some former research\cite{wang2018monte} \cite{hamrick2019combining} has explored the idea of combinations of Q-learning and MCTS, but in simple perfect information games and not applicable to imperfect information games. Our algorithm introduces a more complex MCTS variant with modifications such as a different expansion procedure and alternative backpropagation methods. We have also modified loss functions during agent training based on MCTS returns and developed an MCTS-based reward shaping structure.

In the traditional MCTS, the simulation continues until the game ends after expanding a new state, and a single agent typically plays through the entire game via self-play, with any new state being expanded, evaluated, and backpropagated. However, due to the varying information available to different players in Uno, this approach is not feasible. The player in the new state may be different, and it's impossible for the next player to pass information back to the previous one if the two players are not the same. 

In our approach, We limit each simulation to only one step expansion since this is more computationally efficient. If the simulation starts with player 1’s turn and transitions to player 2’s, we skip evaluating or backpropagating from all states of player 2. Player 2 continues with their own strategy until it's player 1’s turn again. If the game ends in simulation, we backpropagate the results from the end state directly to last state where it was player 1’s turn and along the path to the start state. Our algorithm ensures that all states along the MCTS simulation path belong to the player who starts the simulation.

\begin{figure}[!htbp]
\centering
\includegraphics[width=2.5 in]{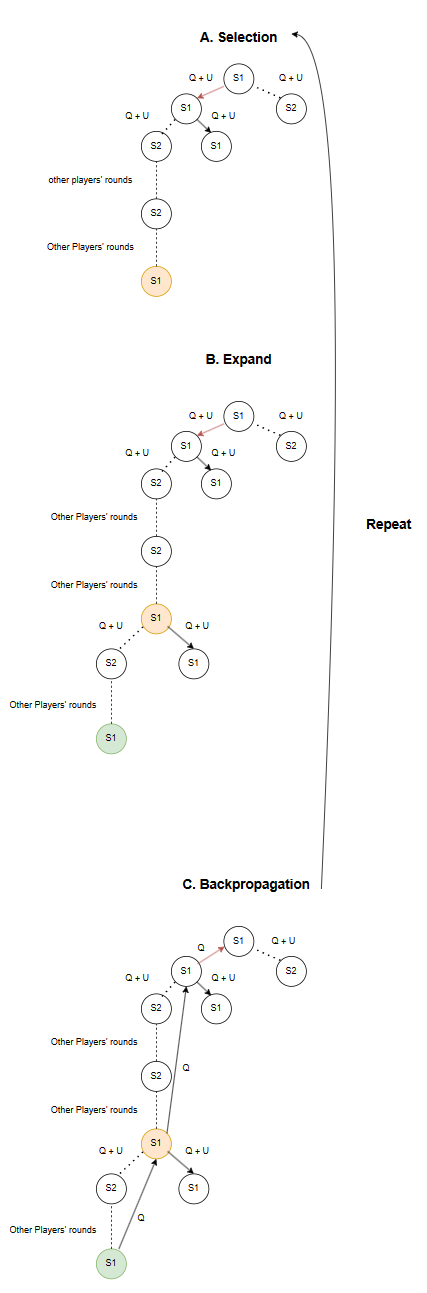}
\caption{The starting player is player 1, so we only expand and backpropagate values of player 1's states.}
\label{mcts}
\end{figure}

The action taken by the agent in expansion is the legal action with highest sum of Q value and the Upper Confidence bounds applied to Trees (UCT):
\begin{align}
\label{q_uct}
\pi(a | s) = \arg \max_a (Q(s, a) + UCT(s, a)),
\end{align}
where $Q(s, a)$ is the Q value of the legal action $a$ under a state $s$, and $UCT(s, a)$ can be defined as:

\[
UCT(s, a) = c_{puct} \sqrt{\frac{N(s)}{1 + N(s, a)}},
\]
where $N(s)$ is the total number of visits of state $s$, and $N(s, a)$ is the total number of uses of legal action $a$ at the state $s$, while $c_{puct}$ is a constant representing the exploration term. When a new state is discovered, if it is not an end state, the Q value $Q(s, a)$ will be predicted for each legal action under that state through neural networks. Q value is passed from the child to the parent during backpropagation and it is updated based on:
\begin{align}
\label{updating_q}
Q_{new}(s, a) = \frac{Q_{old}(s, a) \cdot N(s, a) + Q_{back} (s', a') }{N(s, a) + 1},
\end{align}
where 
\[
Q_{back} (s', a') = \lambda \arg \max_{a'} Q(s', a') + r(s')
\]
$\lambda$ is the reward discount factor. $r(s')$ is the immediate reward after taking $a$ to child state $s'$, always 0 unless the game ends. If $s'$ is the end state, $Q(s', a') $ will be 0, $r(s')$ will be 1 or -1. The whole procedure of our MCTS in Uno follows (Graphical representation shown in Figure~\ref{mcts}):
\begin{enumerate}
    \item \textbf{Selection}: Record the player ID of the root node (start state), $ID_{root}$. This process starts from the root to find a state that has not been expanded. If a state is during player $ID_{root}$'s turn, the agent selects the action with the highest $(Q + UCT)$ based on Equation \ref{q_uct} for simulation. If it is another player’s turn, that player will choose an action based on their own policy until they transition to a state belonging to player $ID_{root}$. 
    \item \textbf{Expansion}: This step unfolds on the previously unexpanded player $ID_{root}$'s state, selecting the action with the highest $(Q + UCT)$ value. We try to find a child, which is also a state of $ID_{root}$, of this unexpanded state. If a game-ending state occurs during another player's turn, we immediately proceed to step 3. If not, the new state of player $ID_{root}$ is found, and this state is initialized by Q value prediction for every legal action, after which we move to step 3.
    \item \textbf{Backpropagation}: This step involves updating the Q values based on Equation \ref{updating_q} for all states of player $ID_{root}$ along the path from the newest found state or end state to the root. Finally, repeat the above three steps for a new round of simulation.
\end{enumerate}

In sampling of standard DDQN, when the agent is at current state $s$, actions are selected via $\epsilon$-greedy: with probability \( \epsilon \), a random action is taken, and with \( 1 - \epsilon \), the action with the highest Q value \( Q(s, a) \) (predicted by neural networks) is chosen. After agent transitioning to the next state \( s' \) and receiving reward \( r \), the tuple \( (s, a, s', r) \) is stored for further training. However, the sampling process in our algorithm is divided into two parts: MCTS simulation and interaction with the real environment. Assume current state \(   s \) in the real environment, which is also the start state of MCTS, is simulated by MCTS. After the simulation, each legal action \( a \) under state \( s \) is assigned a corresponding Q value, \( Q_m(s, a) \), based on our MCTS rules. We then use an $\epsilon$-greedy strategy to select the action based on \( Q_m(s, a) \). Unlike Q values \( Q(s, a) \) are derived from a single neural network in DDQN, our \( Q_m(s, a) \) are obtained from repeated MCTS simulations, reducing overestimations of Q values and making them more accurate by average many Q values backups. During the simulation, whenever an end state is reached and a reward \( r_k \) (-1 or 1) is obtained, we accumulate and average these rewards:
\begin{align}
r_m = \frac{\sum_{k=1}^{n} r_k}{N_s},
\label{eq:reward}
\end{align}
where $n$ is the number of times the reward received at end state, $N_s$ is the number of simulation and $r_m$ is the total average reward from MCTS. After the agent chooses action \( a \) based on \( Q_m(s, a) \) , it transitions to the next state \( s' \) and receives a reward \( r \) from real environment. We then combine \( r \) and \( r_m \) as the total reward $r_t$ after agent takes action \( a \). In standard environment, the agent typically receives reward \( r \) of 0 after taking an action during the ongoing game, but because of rules MCTS, our agent can receive proper rewards at certain points, preventing long periods without positive feedback. We then keep tuple \( (Q_m(s, a), s, a, s', r_t) \) as training data.

We also modified the loss function for training the agent, splitting it into two components:
\begin{align}
\label{loss_function}
L = L_{\text{ddqn}} + L_{\text{mcts}},   
\end{align}
where $L_{\text{ddqn}}$ is the same as in Equation \ref{eq_ddqn1}, except that the $r$ in Equation \ref{eq_ddqn2} is replaced with the total reward $r_t$, and $L_{\text{mcts}}$ is:

\[
L_{mcts} = (Q_{m}(s, a) - Q_{\text{estimator}}(s, a; \theta))^2
\]

Algorithm~\ref{mcts_uno} and \ref{ddqn_mcts} shows the pseudocode of our whole algorithm, including the search and training.

\begin{algorithm}[!htbp]
\caption{MCTS in Uno}
\label{mcts_uno}
\begin{algorithmic}[1]
\Function{INITIALIZE}{}
    \State Initialize data structures of $Q_m(s, a)$, $N(s, a)$, $N(s)$, $V(s)$
    \State $Q_m(s, a)$: set of Q values by taking action $a$ under state $s$
    \State $N(s, a)$: set of numbers of executing action $a$ under state $s$
    \State $N(s)$: set of numbers of visits for state $s$
    \State $V(s)$: set of legal actions for state $s$
    \State $r_k$: rewards get at the end state
    \State Record the player ID of root node as $ID_{\text{root}}$
\EndFunction

\Function{MCTS}{$s$}
    \State \Call{INITIALIZE}{}
    \For{$i \in \{1, \ldots, \text{simulate num}\}$}
        \State $\Call{SIMULATE}{s}$
    \EndFor
    \State $a_{\text{best}} \gets \epsilon-greedy(Q_m(s, a)) $
    \State Calculate $r_m $ based on Equation \ref{eq:reward}
    \State \Return $(Q_m(s, a_{best}), r_m, a_{\text{best}})$
\EndFunction

\Function{SIMULATE}{$s$}
    \If{\Call{ISEND}{$s$}}
        \State Record \Call{REWARD}{s} as $r_k$ \Comment{\Call{REWARD}{s} is the reward function based on the game rules}
        \State \Return 0
    \EndIf
    \If{\Call{ISNEW}{$s$}}
        \State $V(s), Q_m(s, a) \gets \Call{PREDICT}{s}$ \Comment{Predict Q value of every action under state $s$}
        \State $N(s, a) \gets 0$
        \State \Return $\max(Q_m(s, a))$
    \EndIf
    \State $a \gets \arg\max_a \left( Q_m(s, a) + \text{UCT}(s, a) \right)$ \Comment{a in $V(s)$}
    \State $s', r \gets \Call{STEP}{a}$
    \State $Q_m(s', a') \gets \Call{SIMULATE}{s'}$
    \State \Call{UPDATE}{$s$, $a$, $r$, $Q_m(s', a')$}
\EndFunction

\Function{STEP}{$a$}
    \State $s_g \gets \Call{GAMESTEP}{a}$ \Comment{The game environment steps based on rules of game}
    \If{$s_g$ \text{ is state of } $ID_{\text{root}}$} 
        \State \Return $s_g$, $r$
    \EndIf
    \State $a' \gets \Call{POLICYOTHER}{s_g}$
    \State \Return \Call{STEP}{$a'$}
\EndFunction

\Function{UPDATE}{$s$, $a$, $r$, $Q_m(s', a')$}
    \State $N(s, a), N(s) \mathrel{+}= 1$
    \State Update $Q_m(s, a)$ \Comment{Update Q values with formula based on Equation \ref{updating_q}}
\EndFunction
\end{algorithmic}
\end{algorithm}

\begin{algorithm}[!htbp]
\caption{DDQN with MCTS}
\label{ddqn_mcts}
\begin{algorithmic}[1]
\State Initialize experience replay buffer $B$ to keep the training data, the network $Q_{\text{current}}$, and target network $Q_{\text{target}}$
\Function{DDQN}{}
    \For{$i \in \{1, \ldots, \text{training num}\}$}
        \State Initialize the game environment and get the starting state $s$
        \State $B \gets \Call{GENERATEDATA}{s}$
        \State \Call{TRAINING}{B}
    \EndFor
\EndFunction

\Function{GENERATEDATA}{s}
    \While{s is not the end state}
        \State $(Q_m(s, a), r_m, a) \gets \Call{MCTSINUNO}{s}$
        \State $(s', r) \gets \Call{GAMESTEP}{a}$
        \State $r_t \gets r_m + r $
        \State Store the tuple $(Q_m(s, a), r_t, s, a, s')$ into $B$
        \State $s \gets s'$
    \EndWhile
\EndFunction

\Function{TRAINING}{B}
    \State Update the parameters of $Q_{\text{current}}$ using the training data in the buffer $B$ based on Loss Function \ref{loss_function}.
    \If{\Call{ISUPDATETARGET}{}}
        \State Synchronize parameters of $Q_{\text{target}}$ with $Q_{\text{current}}$
    \EndIf
\EndFunction
\end{algorithmic}
\end{algorithm}

\section{Experiments and Evaluations}
We conducted experiments and trained our algorithm DDQN with MCTS, alongside three traditional algorithms—DDQN, DMC, and NFSP—with the same number of training episodes in two Uno environments: a three-player game and a four-player game, configurations commonly played by human players. The performance of DDQN with MCTS was then compared to the three traditional algorithms. To ensure a fair comparison of the performance and learning capabilities of the four algorithms, during training of every reinforcement learning agent, there was only one corresponding RL agent in each environment, while the others were random-playing agents. The evaluation and comparison are based on two main metrics: Total Average Rewards and Win Rate. Rewards were given to the agents at the end of each game by the environment, with the agent receiving either 1 or -1. The total average reward was the sum of rewards accumulated across all games divided by number of games. The win rate was calculated as the number of games won by an agent divided by the total number of games played. Given the high randomness of the Uno game, the reward data exhibits significant variance. To reduce the effect of variance of rewards, after every 1,000 training episodes, the agent was tested by playing 1,000 games with agents playing cards randomly, and the total average rewards were recorded and plotted in the training graph. 

In the evaluation, each algorithm relied solely on its estimator network (DMC, DDQN, DDQN with MCTS) or policy network (NFSP) to select actions. Although DDQN with MCTS used MCTS during the sampling process, for fair comparison, it only used the trained estimator network to select the action with the highest Q value. The code was written in Python 3.10, and agents were trained on a single NVIDIA RTX 4080 GPU. All agents shared the same neural network architecture and hyperparameters, consisting of fully connected layers with sizes 240x64, 64x64, and 64x61. The batch size was set to 32, the learning rate to 0.00005, and the reward discount factor to 0.99. During the sampling process for DDQN with MCTS, each state was simulated 50 times using MCTS. A higher number of simulations could lead to more backpropagation updates and potentially more accurate Q value predictions, but it would also increase the sampling time. We selected 50 simulations as it provides a balance between having rewards $r_m$ based on Equation \ref{eq:reward} in some simulations and avoiding a substantial increase in sampling time. 

\begin{figure*}[!t]
\centering
\subfloat[]{\includegraphics[width=1 in]{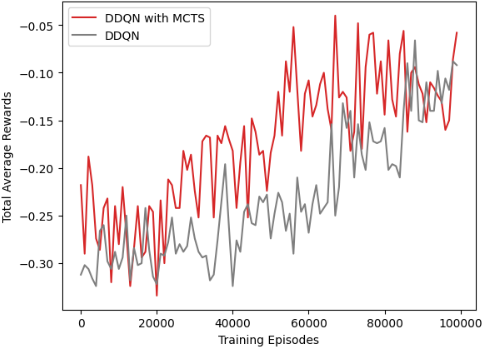}%
\label{mcts_ddqn}}
\hfil
\subfloat[]{\includegraphics[width=1 in]{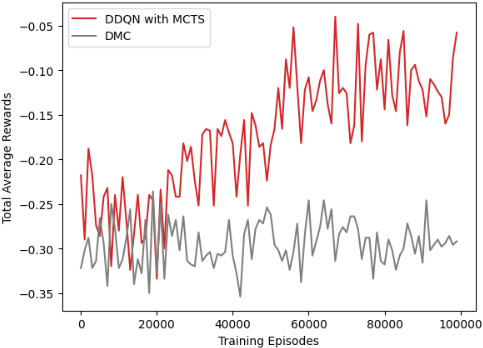}%
\label{mcts_dmc}}
\hfil
\subfloat[]{\includegraphics[width=1 in]{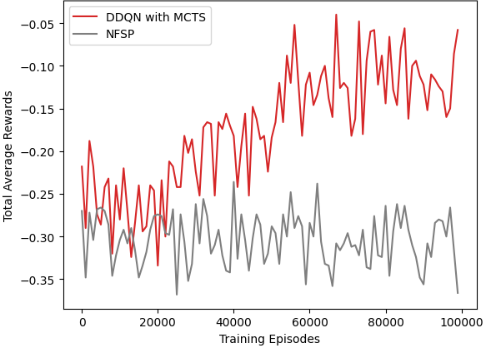}%
\label{mcts_nfsp}}
\hfil
\subfloat[]{\includegraphics[width=1 in]{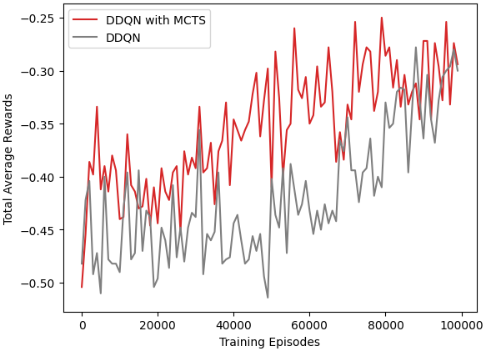}%
\label{mcts_ddqn_4}}
\hfil
\subfloat[]{\includegraphics[width=1 in]{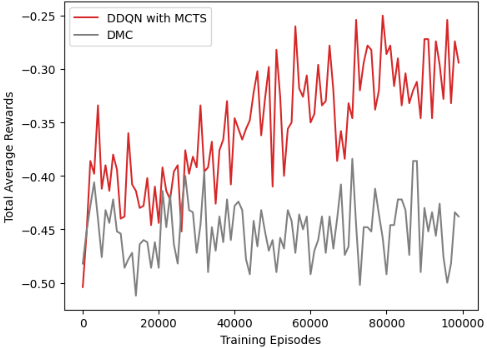}%
\label{mcts_dmc_4}}
\hfil
\subfloat[]{\includegraphics[width=1 in]{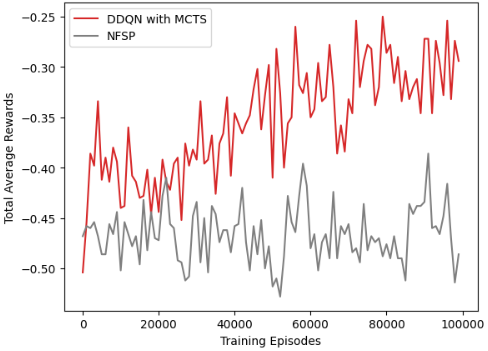}%
\label{mcts_nfsp_4}}
\caption{Comparison of the raw training graphs between DDQN with MCTS  and traditional agents in the 3-player(a, b, c) and 4-player(d, e, f) Uno game. Each data point represents the total average rewards over 1,000 test games where the agent competed against random-playing agents. DDQN with MCTS vs DDQN (a, d). DDQN with MCTS vs DMC (b, e). DDQN with MCTS vs NFSP (c, f).}
\label{fig_Uno_3}
\end{figure*}

\begin{figure*}[!t]
\centering
\subfloat[]{\includegraphics[width=1 in]{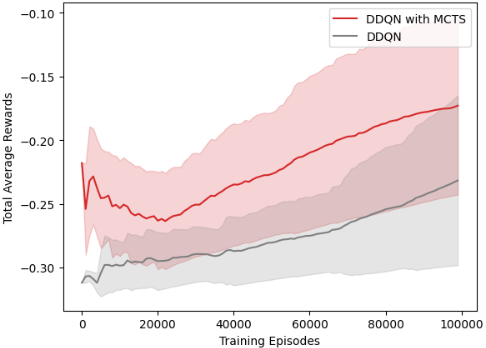}%
\label{mcts_ddqn_clear}}
\hfil
\subfloat[]{\includegraphics[width=1 in]{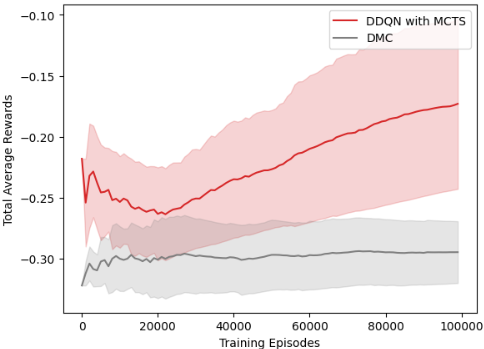}%
\label{mcts_dmc_clear}}
\hfil
\subfloat[]{\includegraphics[width=1 in]{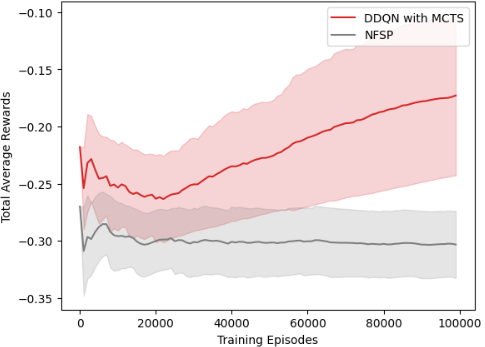}%
\label{mcts_nfsp_clear}}
\hfil
\subfloat[]{\includegraphics[width=1 in]{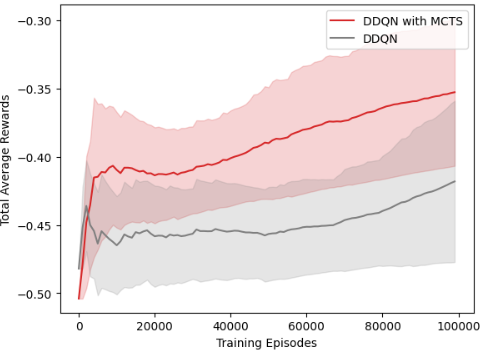}%
\label{mcts_ddqn_4_clear}}
\hfil
\subfloat[]{\includegraphics[width=1 in]{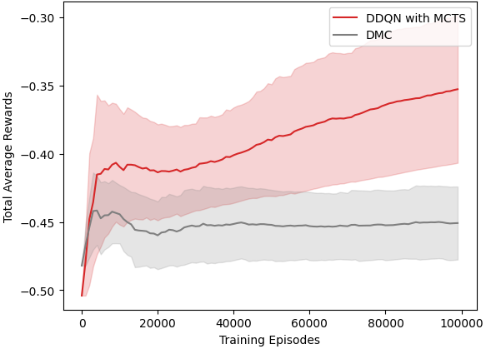}%
\label{mcts_dmc_4_clear}}
\hfil
\subfloat[]{\includegraphics[width=1 in]{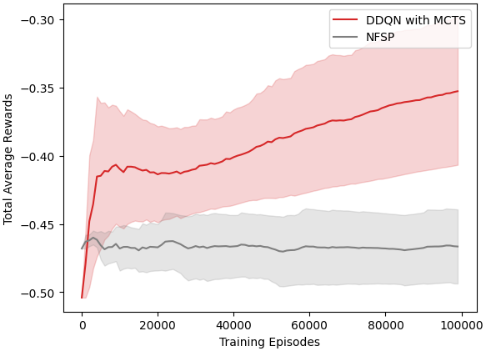}%
\label{mcts_nfsp_4_clear}}
\caption{Comparison of the training graphs(mean and variance) between DDQN with MCTS and traditional agents in the 3-player(a, b, c) and 4-player(d, e, f) Uno game. Each data point represents the total average rewards over 1,000 test games where the agent competed against random-playing agents. DDQN with MCTS vs DDQN (a, d). DDQN with MCTS vs DMC (b, e). DDQN with MCTS vs NFSP (c, f).}
\label{fig_clear_comparison}
\end{figure*}

We trained the four algorithms separately until their performance began to converge. The raw training graphs and comparisons between our algorithm and the other three traditional methods in 3-player and 4-player games are shown in Figure \ref{fig_Uno_3}. DDQN with MCTS achieves the highest average total reward greater than -0.05 in the 3-player game and greater than -0.25 in the 4-player game, indicating a win rate close to 50\% in the 3-player game against random-playing agents, and a win rate higher than 37.5\% in the 4-player game. In general, DDQN with MCTS consistently outperformed the other algorithms in both environments. As the number of players increases, the training difficulty for the agent also grows. While other algorithms tend to perform progressively worse in environments with more players, DDQN with MCTS consistently maintains a stable rate of improvement and learning. More importantly, DDQN with MCTS improves very quickly in the early stages, whereas other algorithms only begin to improve slowly in the middle and later stages. We also focus on the overall trends in the data when comparing these algorithms. Figure \ref{fig_clear_comparison} shows a comparison of the mean and variance of the training data for the algorithms. In terms of the mean comparison, DDQN with MCTS achieves a total average reward that is 0.6 to 1.3 higher than the other algorithms in the 3-player game, meaning its win rate in tests against random-playing agents is 3\% to 6.5\% higher than that of the other algorithms. In the 4-player game, DDQN with MCTS achieves a total average reward that is 0.7 to 1.3 higher, corresponding to a win rate 3.5\% to 6.5\% higher than that of the other algorithms in tests against random-playing agents. We also tested DDQN with MCTS against the other algorithms in 10,000 games to evaluate its win rate when competing directly with them. Regardless of which algorithm it was tested against, DDQN with MCTS always achieved a higher win rate. In the 3-player game, its win rate was 4\% to 16\% higher than that of the other algorithms, and in the 4-player game, it was 5\% to 11\% higher, shown in Table \ref{tab:three_agents} and Table \ref{tab:four_agents}.

\begin{table}[!htbp]
    \centering
    \caption{Win Rate of Four Algorithms in 3-player games test in 10000 games}
    \label{tab:three_agents}
    \begin{tabular}{|c|c|}
        \hline
        Algorithms & Win Rate of Algorithms \\
        \hline
        DDQN with MCTS & \textbf{40\%} $\pm$ \textbf{0.015} \\ 
        DDQN & 35\%  $\pm$ 0.048 \\ 
        DMC & 25\% $\pm$ 0.1 \\ 
        \hline
        DDQN with MCTS & \textbf{40\%} $\pm$ \textbf{0.05} \\ 
        DDQN & 36\% $\pm$ 0.052 \\ 
        NFSP & 24\% $\pm$ 0.005 \\ 
        \hline
        DDQN with MCTS & \textbf{44\%} $\pm$ \textbf{0.048} \\ 
        NFSP & 28\% $\pm$ 0.018 \\ 
        DMC & 28\% $\pm$ 0.05 \\ 
        \hline
    \end{tabular}
\end{table}

\begin{table}[!htbp]
    \centering
    \caption{Win Rate of Four Algorithms in 4-player games test in 10000 games}
    \label{tab:four_agents}
    \begin{tabular}{|c|c|}
        \hline
        Algorithms & Win Rate of Algorithms \\
        \hline
        DDQN with MCTS & \textbf{32\%} $\pm$ \textbf{0.05} \\
        DDQN & 27\% $\pm$ 0.001 \\
        DMC & 20\% $\pm$ 0.1 \\
        NFSP & 21\% $\pm$ 0.015 \\
        \hline
    \end{tabular}
\end{table}
\subsection{Test with Average Human Players and Knowledge learned by DDQN with MCTS}
\begin{figure}[!htbp]
\centering
\includegraphics[width=2.5 in]{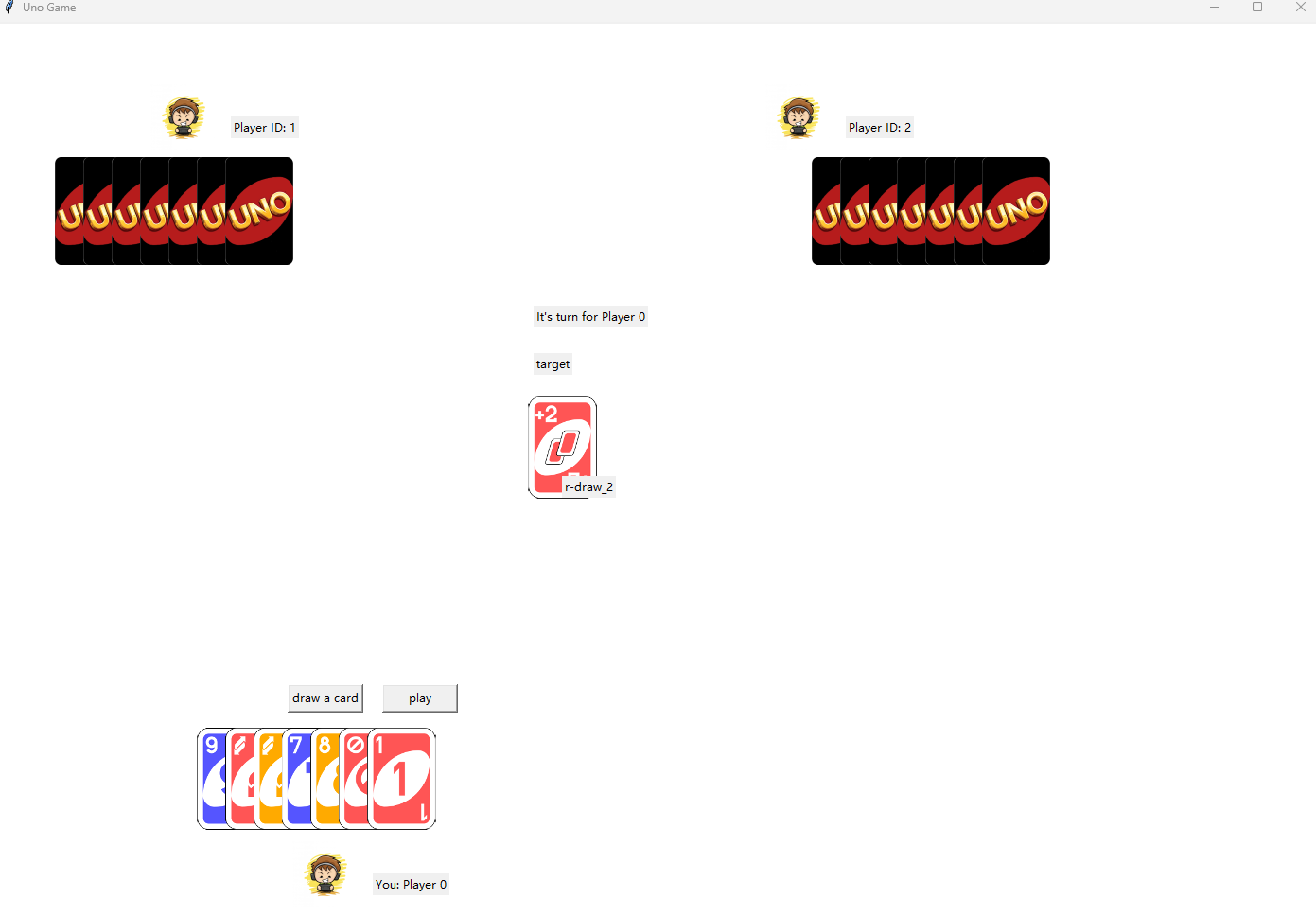}
\caption{Exampled three players GUI}
\label{Uno_2}
\end{figure}

In order to assess the performance of our algorithms in competition with human players, we have developed a fully functional graphical user interface (GUI). Our interface is implemented using the built-in Python module Tkinter \cite{lundh1999introduction}. We are also implementing online multiplayer functionality using the User Datagram Protocol (UDP) \cite{postel1980user}, allowing human players from different locations to remotely compete against our algorithms in 2-players, 3-players and 4-players environments, exampled GUI shown in Figure \ref{Uno_2}.
We conducted a series of 100 games where our algorithm competed one-on-one against ourselves (representing average human players). We observed DDQN with MCTS have surpassed human performance, achieving approximately a 54\% win rate against average human players. In both the 3-player and 4-player games, the win rate of the agent is comparable to that of human players. We also observed that these algorithms indeed learned some patterns of the game, which typically reflect strategies also exhibited by human players. For example, when the agents have many cards along with a 'wild draw 4', they tend to play it immediately. This strategy increases the number of cards the next player holds, making it more challenging for them to win the round. When the agents are down to only two cards, and one of them is a 'wild' or 'wild draw 4', they will save these for the final play. This is because, regardless of the target card’s color or number, the game rules allow a player with a wild card to play it directly, thus creating a guaranteed winning situation by holding onto this card until the end.

\section{Conclusion}
We introduced the complexity of imperfect information games and discussed their research value. We selected Uno, an imperfect information game, as the basis for our research. We represented Uno as a reinforcement learning problem and presented a novel algorithm, Double Deep Q-Learning with Monte Carlo Tree Search (DDQN with MCTS), to address challenges encountered in prior work with imperfect information games, such as reward sparsity and Q value overestimation. Additionally, we developed a graphical user interface (GUI) to allow human players to compete against our agents, where DDQN with MCTS outperformed the average human player in terms of win rates on one-to-one competition. DDQN with MCTS also demonstrated superior performance compared to three traditional methods—Double Deep Q-Learning, Deep Monte Carlo, and Neural Fictitious Self-Play—achieving higher total average rewards with fewer training steps and higher win rates during testing.

In our subsequent work, we plan to experiment with deeper neural networks and more complex network architectures. Due to the long computation time of MCTS, we will also explore optimizing its efficiency \cite{song2022monte}. Furthermore, since our DDQN with MCTS can be generalized and applied to any algorithm's value estimation function, we plan to extend this improvement to state-of-the-art Actor-Critic algorithms \cite{su2021value} \cite{xiao2022asynchronous}.

\section*{Acknowledgments}
We would like to thank Amitabh Trehan for his advice in the development of this research.

\bibliography{references}

\end{document}